\documentclass[letterpaper, 10 pt, conference]{ieeeconf}  % Comment this line out if you need a4paper

\IEEEoverridecommandlockouts                              % This command is only needed if 
\usepackage{graphics} % for pdf, bitmapped graphics files
\usepackage{epsfig} % for postscript graphics files
\usepackage{mathptmx} % assumes new font selection scheme installed
\usepackage{times} % assumes new font selection scheme installed
\usepackage{amsmath} % assumes amsmath package installed
\usepackage{amssymb}  % assumes amsmath package installed
\usepackage[pdfborder={0 0 0}, colorlinks=false]{hyperref}
\usepackage{graphicx}

\title{\LARGE \bf
Optimizing Grasping in Legged Robots: A Deep Learning Approach to Loco-Manipulation
}

\author{Dilermando Almeida$^{1}$, Guilherme Lazzarini$^{1}$, Juliano Negri$^{2}$, \\
Thiago H. Segreto$^{2}$, Ricardo V. Godoy$^{2,*}$ and Marcelo Becker$^{2}$% <-this % stops a space
\thanks{
This work was supported by the Petr\'{o}leo Brasileiro S/A - Petrobras,
using resources from the R\&D clause of the ANP, in partnership with the
Universidade de S\~{a}o Paulo (USP) and the Funda\c{c}\~{a}o de Apoio \`{a} F\'{\i}sica e
\`{a} Qu\'{\i}mica (FAFQ), under Cooperation Agreement No. 2023/00016-6 and
2023/00013-7.}
\thanks{$^{1}$Dilermando Almeida, Guilherme Lazzarini are with the College of Mechanical Engineering, Federal University of Uberlândia, Uberlândia, Brazil
        {\tt\small dilermandoada@gmail.com}}%
\thanks{$^{2}$Juliano D. Negri, Thiago H. Segreto, Ricardo V. Godoy, and Marcelo Becker are with the  Department of Mechanical Engineering, University of São Paulo, São Carlos, Brazil.}
\thanks{$^{*}$Corresponding author: {\tt\small ricardo.godoy@alumni.usp.br}}}%

\begin{document}

\maketitle
\thispagestyle{empty}
\pagestyle{empty}

%%%%%%%%%%%%%%%%%%%%%%%%%%%%%%%%%%%%%%%%%%%%%%%%%%%%%%%%%%%%%%%%%%%%%%%%%%%%%%%%
\begin{abstract}

This paper presents a deep learning framework designed to enhance the grasping capabilities of quadrupeds equipped with arms, with a focus on improving precision and adaptability. Our approach centers on a sim-to-real methodology that minimizes reliance on physical data collection. We developed a pipeline within the Genesis simulation environment to generate a synthetic dataset of grasp attempts on common objects. By simulating thousands of interactions from various perspectives, we created pixel-wise annotated grasp-quality maps to serve as the ground truth for our model. This dataset was used to train a custom CNN with a U-Net-like architecture that processes multi-modal input from an onboard RGB and depth cameras, including RGB images, depth maps, segmentation masks, and surface normal maps. The trained model outputs a grasp-quality heatmap to identify the optimal grasp point. We validated the complete framework on a four-legged robot. The system successfully executed a full loco-manipulation task: autonomously navigating to a target object, perceiving it with its sensors, predicting the optimal grasp pose using our model, and performing a precise grasp. This work proves that leveraging simulated training with advanced sensing offers a scalable and effective solution for object handling.

%%%%%%%%%%%%%%%%%%%%%%%%%%%%%%%%%%%%%%%%%%%%%%%%%

\end{abstract}

\section{Introduction} \label{sec:introduction}
Quadruped robots have emerged as highly efficient and versatile platforms, excelling in navigating complex and unstructured terrains where traditional wheeled robots might fail. Equipping these robots with manipulator arms unlocks the advanced capability of loco-manipulation, enabling them to perform complex physical interaction tasks in areas ranging from industrial automation to search-and-rescue missions. However, achieving precise and adaptable grasping in such dynamic scenarios remains a significant challenge, often hindered by the need for extensive real-world calibration and pre-programmed grasp configurations. The capability of locomotion, coupled with the aptitude for performing elaborate missions, including the creation of spatial models, item identification, and physical contact with the environment, drives the recent rise of quadruped robots in the area of automated systems~\cite{Papadopoulos2023Legged, Solmaz2024Robust}. When equipped with robotic arms and grippers, these robots combine locomotion and manipulation, a concept known as loco-manipulation, enabling interactions with the environment that would not be possible with stationary or wheeled robots~\cite{Papadopoulos2023Legged}. This capability is especially valuable in real-world applications, such as industrial automation, search and rescue missions, and assistive technologies, where robots must operate in dynamic and unstructured environments~\cite{Solmaz2024Robust}.

Loco-manipulation has been studied and discussed for decades, with initial research focused on the development of stable locomotion systems~\cite{Solmaz2024Robust} and the rudimentary integration of end-effectors for manipulation~\cite{MachadoSilvaLeggedRobots}. In the pioneering stages, the emphasis was on rigid kinematic models and control strategies based on fixed rules, which guaranteed basic functionality in controlled environments~\cite{MachadoSilvaLeggedRobots}. Over time, advances in high-precision sensors, computer vision techniques, and motion planning architectures have enabled greater autonomy and adaptability, particularly on irregular terrain. However, even with the introduction of optimization algorithms and reinforcement learning, the coordination between locomotion and manipulation in unpredictable scenarios continued to require complex and customized solutions, often linked to specific contexts~\cite{Liu2024Overview}.

Despite advances, achieving precise and adaptable manipulation in such scenarios remains a significant challenge due to the need for robust object recognition, accurate grasp planning, and fluid integration with the locomotion system~\cite{Guo2024FrameworkGrasp, Mahler2018DeepLearningGrasping}. Traditional approaches rely on predefined grasp configurations or intensive manual calibration, which limits their flexibility~\cite{Guo2024FrameworkGrasp, Mahler2018DeepLearningGrasping}. Alternatively, the use of deep learning has shown promise by allowing robots to learn generalized grasping strategies from data, overcoming these limitations~\cite{morrisonclosing2018, Guo2024FrameworkGrasp, Mahler2018DeepLearningGrasping}.

Inspired by recent works~\cite{kasaei2022simultaneousmultiviewobjectrecognition, morrisonclosing2018}, this work proposes a new deep learning-based framework specifically designed for quadruped robots. The goal is to optimize picking tasks by training a model that can recognize objects and determine optimal grasping points from RGB-D data. The majority of machine learning (ML) algorithms used for object grasping return the end-effector opening, its orientation, and grasp quality~\cite{morrisonclosing2018, Guo2024FrameworkGrasp}. Inspired by these approaches, we proposed a pipeline that integrates a pre-trained object detection model, a supervised learning model that classifies optimal grasping points, and a normal estimation model. To achieve this, a custom dataset was created from simulated environments using the Genesis platform where the robot's end-effector could interacts with objects representing everyday items, capturing RGB-D information from multiple views, and generating grasp annotations for different shapes.

This simulated dataset was used to train a convolutional neural network (CNN) capable of using depth camera outputs as inputs to map optimal grasping points. This approach allows for handling a wide variety of grasping scenarios without relying on extensive real-world data collection. After training, the model was implemented on for performing elaborate missions, including the creation of

\begin{enumerate}
    %\item The development of a simulation-based training pipeline, which reduces dependence on data collected in the real world;

    \item The development of a simulation-based training pipeline on Genesis Simulator, which allows for massive parallel data collection and doesn't require real data;
    
    \item The incorporation of advanced perception methods, such as D2NT, to improve the precision of depth-based grasping and reduce computational cost related to ML execution;

    \item The development of flexible frameworks capable of integrating with high-level control APIs and commercial robots that lack low-level access;
    
    \item The validation of the methodology on a physical robot, demonstrating the effectiveness of the approach in real-world scenarios.
\end{enumerate}

This research poses an important advancement in performing manipulation tasks with quadruped robots by integrating deep learning, advanced simulation, and robotic control. The proposed approach paves the way for the development of autonomous, versatile, and intelligent systems capable of operating effectively in unpredictable environments with optimized processing speed.

\section{Dataset Extraction} \label{sec:datasetextraction}
The dataset employed in this work to train the proposed deep learning model was entirely generated in a simulated environment from successive grasping attempts. For this, we used the Genesis framework, a simulation platform designed for physical AI/robotics/Embodied and general-purpose applications~\cite{Genesis}. Its main advantages are the ability to simulate parallel environments efficiently and the availability of RGB and depth cameras.

To serve as a grasping target, a 3D model of a water bottle~\cite{Free3DWaterBottle} was selected. To evaluate grasping, only the end-effector of the robotic arm, attached to the quadruped, was used, as it must align and make contact with the object.

\begin{figure}[h!]
    \centering
    \includegraphics[angle=0, width=5cm, trim={15cm 4cm 15cm 5cm},clip]{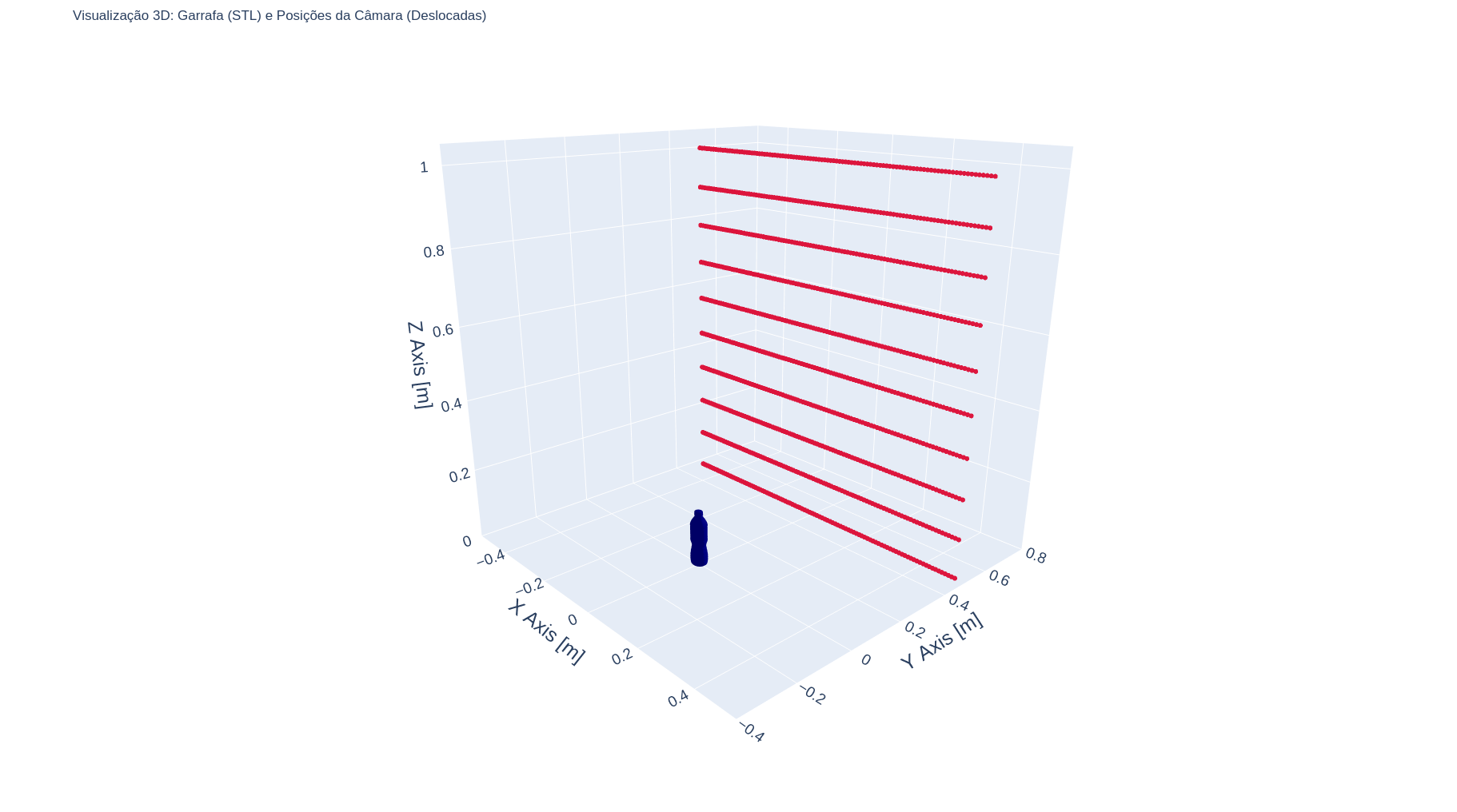}
    \caption{Illustration of the camera positioning process used. The water bottle is shown at the origin, in blue, and the camera points are represented as red dots, located at y=0.5m, as the x and z position varies between -0.5m and 0.5m.}
    \label{Fig:cam_points}
\end{figure}

The first step to generate the dataset was to configure a virtual RGB-D camera to extract images of the object. For this, we positioned the items to be picked at a fixed point and sampled images from positions on a 2D grid composed of 100 points on the X-axis and 10 points on the Z-axis (both between -0.5m and 0.5m), with the Y-axis fixed at $y~=~0.5m$. This resulted in 1000 distinct positions, where the cameras were positioned, such that each of them looked at a slightly random point in the X and Y coordinates (between -0.03m and 0.03m), and Z (between 0m and 0.09m). Figure~\ref{Fig:cam_points} shows the camera position points used. In this way, we generated a thousand different views, with each view containing a dataset with RGB, depth, segmentation masks, and normal map information.

% Such a dataset was used as input for the neural network and could be extracted with the aid of the simulator's own API, in Python. As for the target, ten thousand parallel simulations were performed for each view of grasping attempts.

\begin{figure}[h!]
    \centering
        \includegraphics[angle=0, width=5cm]{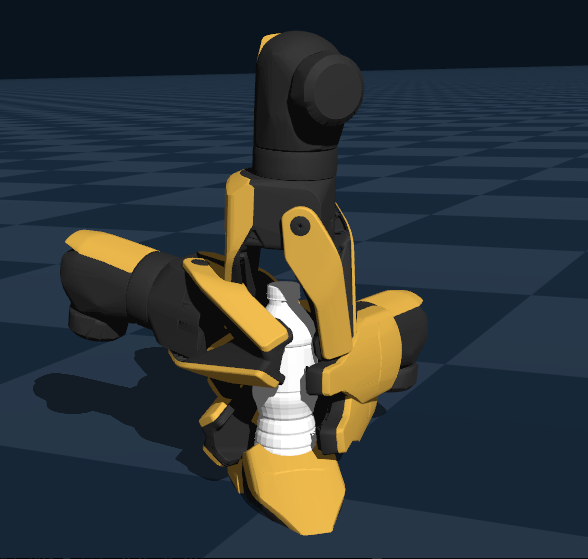}
    %\caption{Parallel simulation for grasping, from \cite{sei_la_oq_sei_la_oq}}
    %\caption{Parallel grasping simulation performed in the Genesis World environment, illustrating five robotic grippers executing simultaneous grasping attempts on a geometric model of a water bottle. Each gripper is aligned with a specific pixel of the object's image from the normal at that point, testing contact with the bottle's surface and the absence of contact with the ground, according to the Boolean logic of the simulation pipeline.}
    \caption{Parallel grasping simulation performed in the Genesis World environment, illustrating five robotic grippers executing simultaneous grasping attempts on a geometric model of a water bottle. Each gripper is aligned with a specific pixel of the object's image from the normal at that point.}
    \label{Fig:5}
\end{figure}

The second step is to generate grasping poses for each pixel of the object in each image and label them as successful or failed. To achieve this, we leveraged the simulator segmentation mask collected by the camera, and for each selected pixel in the object's segmentation mask, an end-effector was assigned to attempt to grasp it. The coordinate of the pixel under analysis was converted into the global frame coordinates of the simulated environment, along with the normal associated with it. For each grasping pose, the end-effector was positioned 1.0 m from the object to be picked up, and its initial orientation was parallel to the ground, pointing towards the object; then the end-effector attempted to grasp at a distance of 0.35 m from the surface of the object and aligned with the respective normal vector, in relation to the gripper's internal frame. Those parameters were experimentally chosen to ensure the end-effector could properly align with the point on the surface of the given object. The final position of the gripper is shown in Fig.~\ref{Fig:5}.

\begin{figure}[h!]
    \centering
    \includegraphics[angle=0, width=5cm]{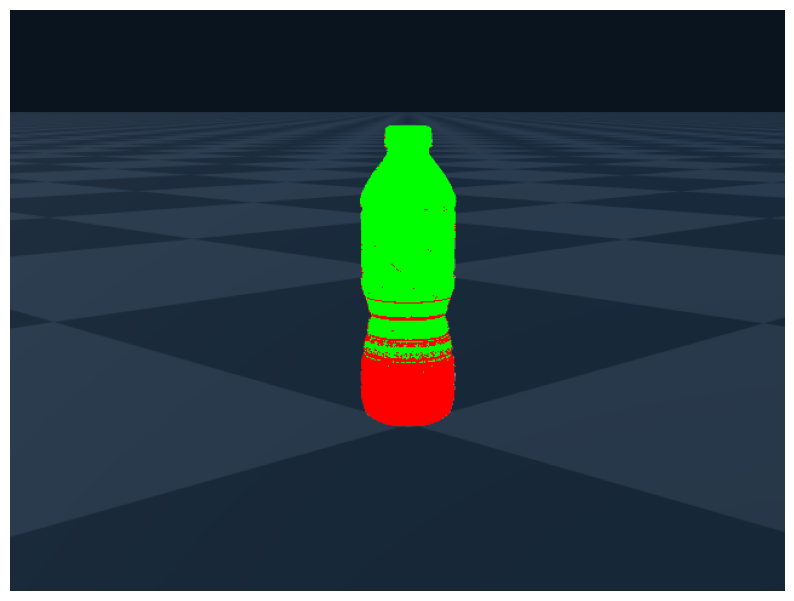}
    %\caption{Dataset Ground Truth, from \cite{sei_la_oq_sei_la_oq}}
    \caption{Representation of the dataset ground truth for mapping grasping points, generated from simulations in the Genesis World environment. Pixels encoded in green indicate successful grasping regions, and pixels in red indicate where grasping failed. Pixels outside the segmentation mask are considered indeterminate.}
    \label{Fig:4}
\end{figure}

To map successful grasping poses, we relied on the contact information of the end effector with the object. Therefore, we selected the collision geometries of the gripper of the robotic manipulator to extract the contact information with the object's collision geometry. The grasp pose was considered successful (True) if all given points in the gripper collided with the bottle without any other external collision (e.g. contact with the ground), or failed (False) otherwise. From this, a grasping mask was generated with pixel values as follows: those that met the positive grasp criteria were marked with value 1, those that did not meet the criteria were marked with value 0, and pixels outside the object's segmentation mask, i.e., regions where no object was present, were considered indeterminate, receiving the value -1. The result of this process can be visualized in Fig.~\ref{Fig:4}, where pixels in green indicate successful grasps and those in red indicate unsuccessful grasps.

% \section{Experiments} \label{sec:experiments}
\section{Experimental Results} \label{sec:experimental_results}
% \subsection{Machine Learning Framework}

\begin{figure*}[t]
    \centering
    \includegraphics[width=\textwidth]{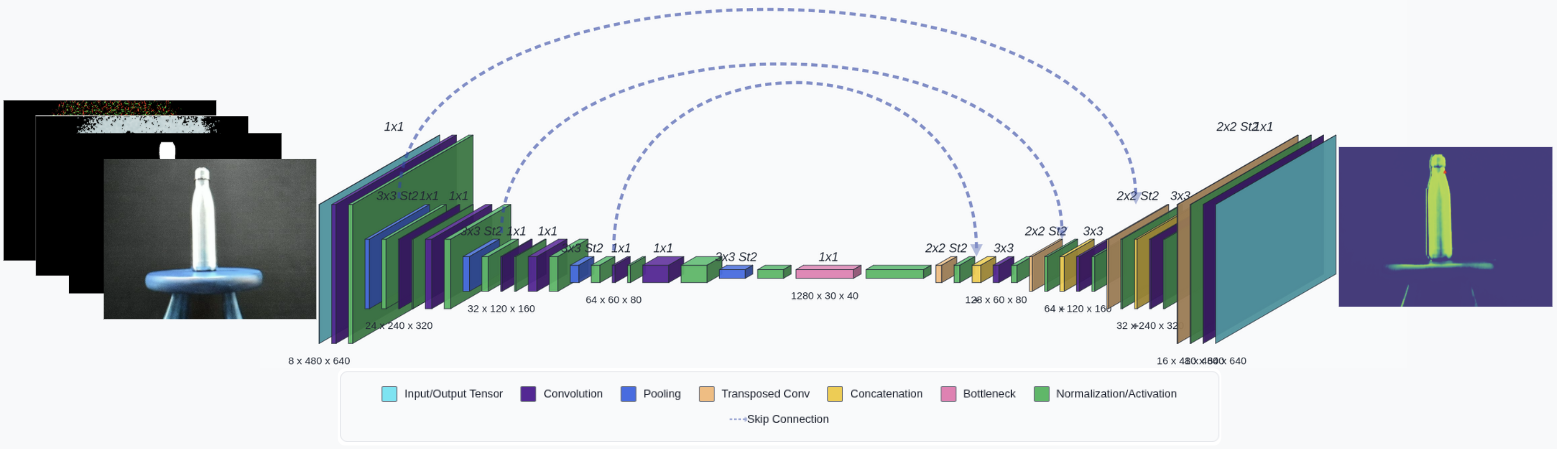}
    %\caption{Neural Network architecture used }
    \caption{Model used to predict optimal grasping points. The inputs (normal map, depth, segmentation, and RGB image) (left) are processed by the CNN to generate a map of optimal grasping points (right), indicated by the light green region. The optimal point, indicated by the red dot in the output image, is determined by the network.}
    \label{Fig:nn_arch}
\end{figure*}

An ML framework is designed to identify optimal grasp points by processing multi-channel sensor data from an RGB-D camera. We implemented a fully convolutional encoder-decoder architecture, similar to U-Net~\cite{ronneberger2015unetconvolutionalnetworksbiomedical}, to perform semantic segmentation on 480x640x8 inputs (RGB, depth, normal map, and segmentation mask), as shown in Fig.~\ref{Fig:nn_arch}. The encoder, derived from MobileNetV2~\cite{kerasmobilenet}, systematically extracts features while reducing spatial resolution. The decoder reconstructs the image, using transposed convolutions for upsampling and skip connections to merge fine-grained features from the encoder, which is critical for precise localization. The model, which contains approximately 5.44 million trainable parameters, uses \texttt{GroupNorm} in its decoder for improved training stability. A final 1x1 convolution produces a single-channel output map where the pixel with the highest value indicates the best-predicted point for grasping.

\begin{figure}[h!]
\centering
\includegraphics[angle=0, width=8cm, trim=0 2px 0 0, clip]{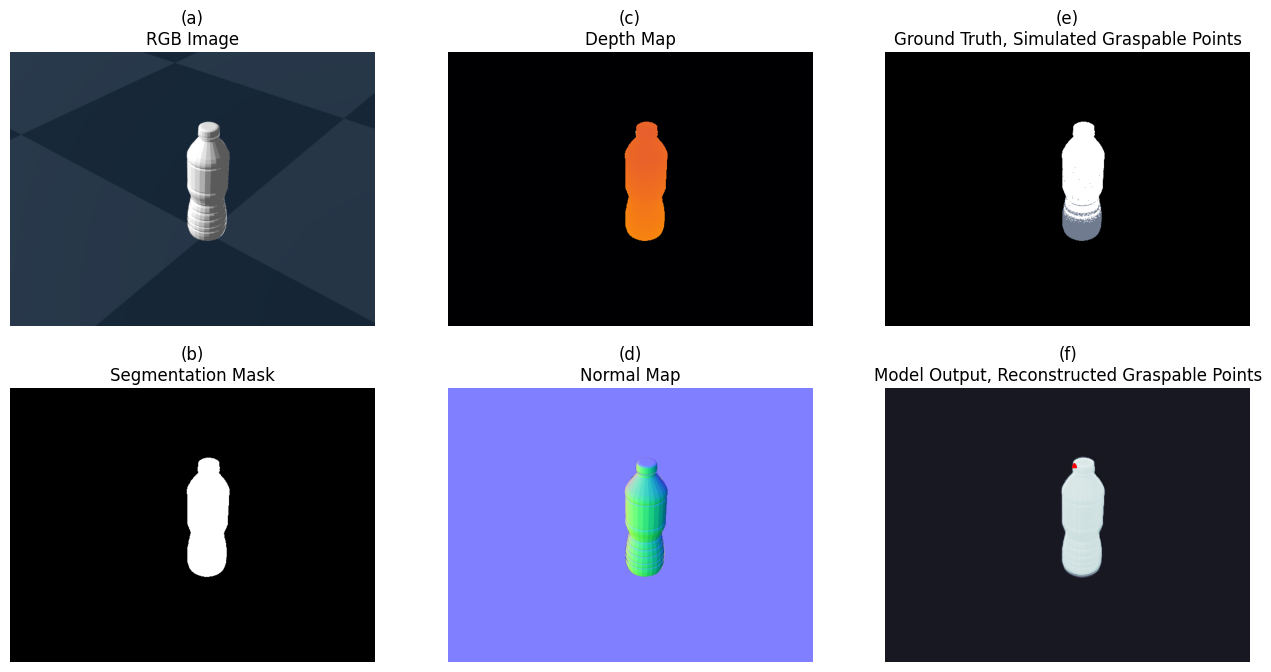}
\caption{Visualization of input data extracted in the Genesis World ((a) RGB image, (b) segmentation mask,  (c) depth map (d) normal map, (e) ground truth of viable pixels for grasping (white for success, gray for failure, and black for no data)) and the corresponding neural network output for those inputs (f)indicating the probability of grasping success per pixel (100\% for white, 0\% for black).}
\label{Fig:0}
\end{figure}

The model was trained on a synthetic dataset generated entirely within the Genesis simulation environment. This data, captured by a virtual RGB-D camera at a 480x640 resolution, mirrors the specifications of the physical sensor. The corresponding ground truth for training is a grasp-quality map of the same resolution, where each pixel is labeled based on the outcome of simulated grasp attempts: 1 for a successful grasp, 0 for a failed grasp, and -1 for indeterminate regions, such as the background. All the inputs, ground truth, and model output can be visualized in Fig.~\ref{Fig:0}.

% \section{Results} \label{sec:results}
\section{Deployment}
The development of robust robotic systems for grasping is fundamental to advancing autonomous manipulation in unstructured environments. This section describes the deployment of a pipeline for object grasping, utilizing the Boston Dynamics Spot platform, which enables precise and adaptive movement of a generic thermos bottle as a test object. The following subsections describe the robotic platform, experimental setup, execution pipeline, grasp point identification process, and grasp execution strategy, providing a comprehensive overview of the system's implementation.

\subsection{Robotic Platform}
\begin{figure}[h!]
\centering
\includegraphics[angle=0, width=8.5 cm]{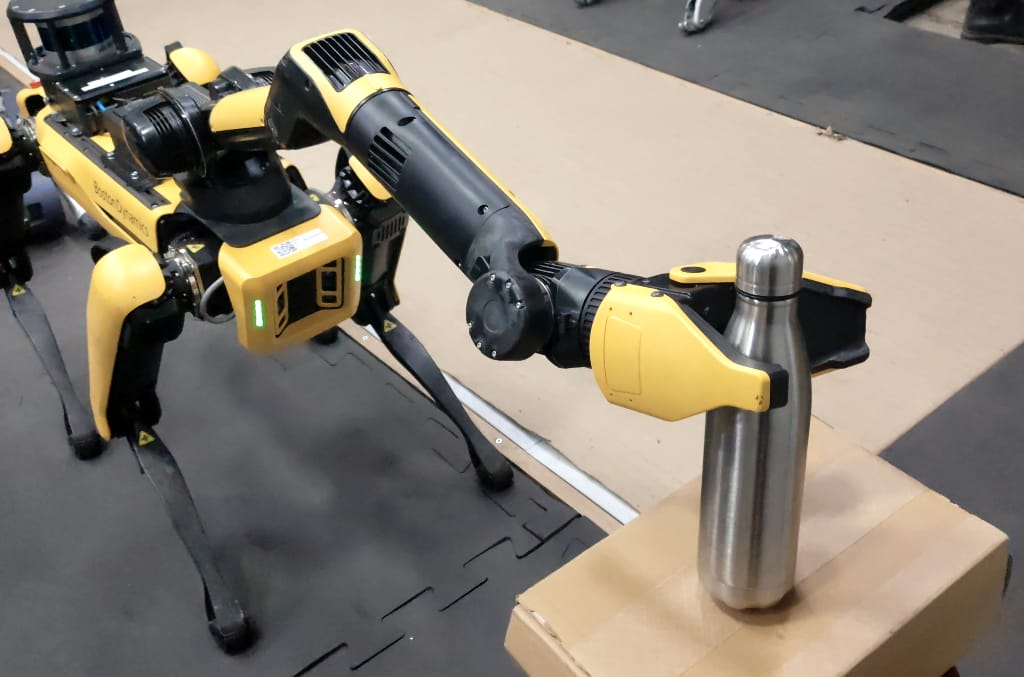}
%\caption{Robot system grasping thermal bottle, from \cite{sei_la_oq_sei_la_oq}}
\caption{Photo of the quadruped robot during the deployment of the grasping pipeline in a real-world scenario, capturing a thermos bottle. The manipulator arm is designed so that the gripper aligns with the point selected by the neural network.}
\label{Fig:3}
\end{figure}

The robotic platform used in this work was the Boston Dynamics Spot, a quadrupedal robot. Equipped with a robotic manipulator, Spot offers a versatile system capable of navigating complex terrains while performing precise manipulation tasks. The manipulator, integrated into Spot's chassis, provides six degrees of freedom, facilitating its interaction with objects. The manipulator also includes a jaw gripper attached to its end-effector. For this study, a generic thermos bottle was selected as the target object, representing a common household item with a cylindrical geometry and a smooth surface, which is suitable for evaluating grasping performance. The entire setup can be seen in Fig.~\ref{Fig:3}.

\subsection{Experimental Setup}
The experimental deployment was conducted in a real-world environment to validate the grasping pipeline. The Boston Dynamics Software Development Kit (SDK)~\cite{BostonDynamicsSpotSDK} was utilized to control the quadrupedal-manipulator system, providing programmatic access to locomotion, manipulation, and sensor data. Spot's onboard sensor suite, such as RGB cameras and depth sensors mounted on the end-effector, facilitated comprehensive data acquisition. These sensors captured essential visual and spatial information for object detection and grasp planning.

Object recognition and segmentation were performed using the pre-trained vision model \textit{YOLOv11}~\cite{yolo11_ultralytics}, a state-of-the-art CNN optimized for real-time performance. The model was fine-tuned to detect and segment objects, including bottles, generating a bounding box as well as precise binary masks that delineated the object's contours in the camera's field of view. The integration of these components, SDK control, sensor data, and vision model, formed a robust foundation for the grasping experiments, ensuring reliable perception and manipulation in real-world conditions.

\subsection{Overview of Execution Steps}
The grasping pipeline was executed through a sequence of steps, designed to transition from simulation to physical deployment while building a dataset for model training. Importing Spot's manipulator end-effector into the Genesis environment allowed preliminary testing of the grasping pipeline in a controlled virtual setting. With this, grasping attempts were executed, involving the identification and manipulation of the bottle. Data extracted from these attempts, including RGB images, depth maps, and grasping outcomes, were aggregated to construct a dataset for training a neural network model tailored for grasp point prediction.

The model was trained with this dataset, utilizing supervised learning to map sensor inputs to viable grasp configurations. Prior to physical deployment, the complete pipeline was rigorously tested in the simulation to validate its performance and mitigate risks associated with hardware operation. Successful simulation results justified the transition to the physical Spot robot, where the pipeline was executed under real-world conditions. This iterative process ensured robustness and reliability in both virtual and physical domains.

\subsection{Grasping Pipeline}

\begin{figure*}[t]
\centering
\includegraphics[width=\textwidth, trim=0 7.5cm 0 7.5cm, clip]{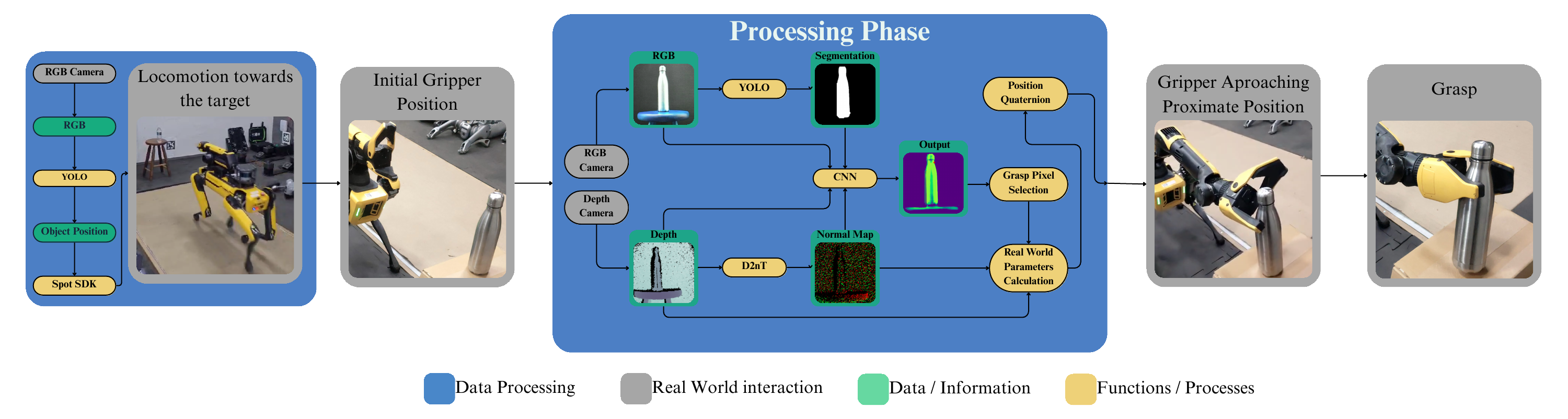}
\caption{Proposed pipeline scheme, beginning with the robot finding and walking towards the object, then initializing RGB-D data acquisition. Then there is parallel processing for object segmentation and normal map generation, and prediction of the grasp point, leading to grasping parameters calculation, and the grasping action execution.}
\label{Fig:deploy_pipeline}
\end{figure*}

The grasping pipeline is executed through a carefully planned sequence of steps, each designed to process sensor data and generate precise grasping parameters, as shown in Fig.~\ref{Fig:deploy_pipeline}. First, the quadruped needs to recognize the object within the area, then walk towards a proximate position, 1.0 m away from the object, enabling the end-effector camera to work as intended. Then, the manipulator's gripper is initialized in an open state, preparing it for interaction with the object. This ensures that the gripper is optimally positioned to capture unobstructed RGB-D data from its external cameras, maximizing the field of view and enabling clear data acquisition. Subsequently, the RGB and depth cameras embedded in the gripper capture a synchronized pair of images, providing comprehensive visual and spatial information about the environment. The RGB image and the depth map, both with a resolution of 480×640 pixels, are obtained using the \textit{hand color image} and \textit{hand depth in hand color frame} sources, respectively, serving as primary inputs for subsequent processing stages.

Subsequently, the \textit{YOLOv11} model processes the RGB image to detect and segment the thermos bottle, generating a binary mask that effectively isolates the object from the background. To ensure compatibility with subsequent processing, this mask is resized to match the depth map's resolution, maintaining consistency across different data modalities. The depth map is then transformed into a normal map using the \textit{d2nt} algorithm, which estimates surface normals based on depth gradients. This conversion relies on approximate camera intrinsics, with focal lengths ($f_x, f_y \approx 554.26$) pixels and a principal point at ($u_0 = 320, v_0 = 240$), calculated for the resolution of the gripper's embedded cameras, to compute 3D point clouds and normals essential for planning the gripper's orientation during grasping.

\begin{figure}[h!]
\centering
\includegraphics[angle=0, width=8cm, trim=0 3px 0 0, clip]{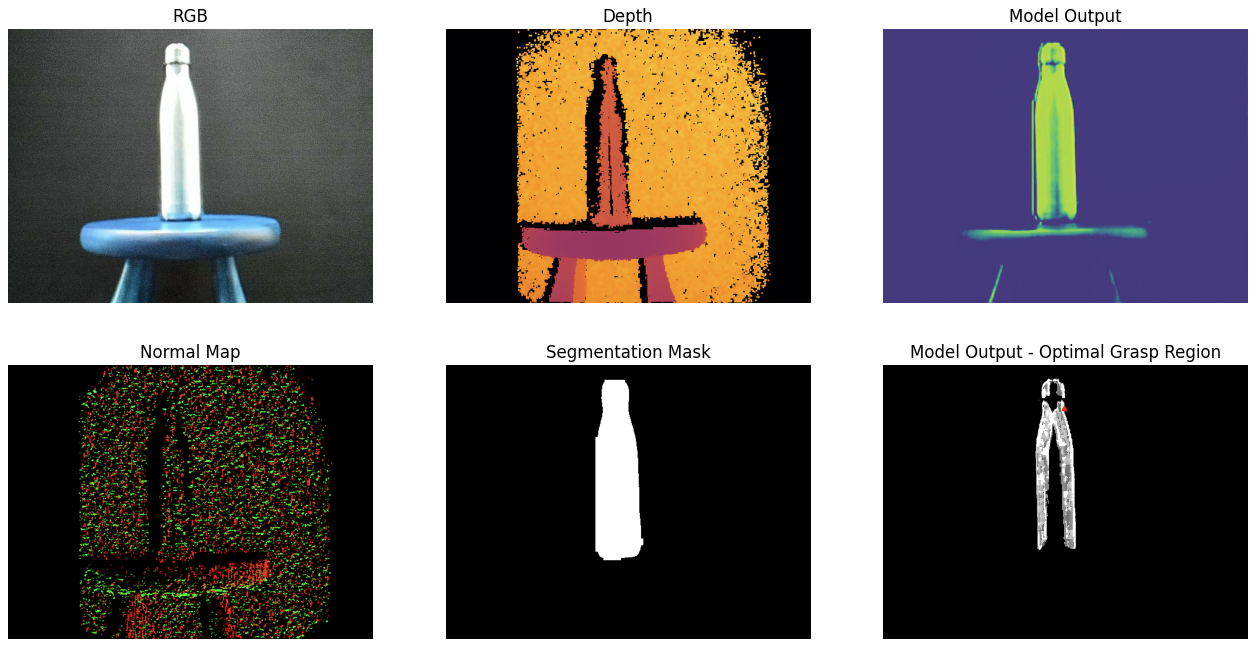}
%\caption{Neural Network output using real data as input, from \cite{sei_la_oq_sei_la_oq}}
\caption{Processed input data captured by the gripper's RGB-D cameras in a real-world scenario ((a) RGB image, (b) depth map, (c) segmentation mask by YOLO-11, (d) normal map generated by the d2nt) and grasp point prediction on a thermos bottle((e) grasping chance heat map, (f) optimal grasping region, each point deviates up to 15\% from 1).}
\label{Fig:1}
\end{figure}

Next, the RGB image, depth map, normal map, and segmentation mask undergo pre-processing to align with the model's input requirements. This involves resizing the RGB image and mask to 480×640, normalizing pixel intensities (RGB and depth to $[0, 1]$), and ensuring normal vectors have unit length. This pre-processing reduces noise and ensures data consistency, preparing the inputs for the next stage. The pre-processed data is then fed to the CNN, which is trained to predict optimal grasping points. The network generates a Boolean map indicating success and failure pixels for grasping, constrained by the segmentation mask to focus exclusively on the object, as displayed in~Fig.~\ref{Fig:1}.

From this map, a pixel is selected as the grasping point. Using the depth value and surface normal at this pixel, the system calculates 3D coordinates and orientation. The position is determined by equations Eq.~(\ref{Eq:1}), Eq.~(\ref{Eq:2}), and Eq.~(\ref{Eq:3}), where $(u, v)$ represents the pixel location. The normal vector is converted to Euler units, and only then is it converted to Quaternion to facilitate manipulator control. Finally, the calculated grasping parameters are translated into a manipulator command via the Boston Dynamics SDK, directing the gripper to securely close around the thermos bottle, completing the grasping action.
\begin{equation}
x = \frac{(u - u_0)z}{f_x},
\label{Eq:1}
\end{equation}

\begin{equation}
y = \frac{(v - v_0)z}{f_y},
\label{Eq:2}
\end{equation}

\begin{equation}
z = \text{depth},
\label{Eq:3}
\end{equation}

\subsection{Locomotion, Grasping Execution and Control}

Locomotion consisted of having the quadruped autonomously approach the identified object, while grasping execution involved positioning the manipulator gripper at the 3D coordinates calculated from the pixel selected by the CNN and orienting it according to the specified quaternion. The thermos bottle, with its cylindrical shape and smooth surface, was grasped by closing the gripper to securely envelop the object, ensuring contact force without deformation. Force-limited control was implemented through the SDK, which enabled force detection upon end-effector contact.
This approach enabled the gripper to stop applying additional force upon detecting resistance from the object, adapt to external forces, and maintain a stable grip. With a maximum torque of 3.0 Nm, the system mitigated the risk of slippage or excessive force application, accommodating minor misalignment or irregularities on the bottle's surface. The successful execution of grasps highlights the effectiveness of this control strategy. The entire project is available at the Mobile Robotics Group's GitHub repository at:

\begin{center}
\href{https://github.com/EESC-LabRoM/dl_grasping_loco_manipulation_2025}{Mobile Robotics Group's Github}
\end{center}

\section{Conclusion} \label{sec:conclusion}
% This study presented an innovative approach to optimize grasping in quadruped robots, integrating deep learning and locomotive manipulation into a pipeline implemented on the Spot platform. Through the construction of a detailed dataset, the training of a custom neural network, and the deployment of a vision-guided grasping system, the project achieved significant contributions to the field of quadruped robotics. The ability to accurately detect, segment, and manipulate objects, utilizing the \textit{YOLOv11} model for segmentation and the \textit{Depth to Normal Translator} method for surface normal mapping, demonstrated the feasibility of combining advanced perception with robot-world interaction in real-world scenarios.
This study presents an innovative pipeline for locomotive manipulation in quadruped robots, integrating a deep learning model for object segmentation with a method for estimating surface normals. Through the construction of a detailed dataset, training a custom neural network, and deploying a vision-guided grasping system, this work made significant contributions to the field of quadruped robotics. The ability to accurately detect, segment, and manipulate objects, demonstrates the feasibility of combining advanced perception with robot-world interaction in real-world scenarios.

Despite these advancements, the project faced notable challenges that require ongoing attention. The generalization of the ML model was limited by variations in object geometry and texture, restricting its effectiveness to objects with specific characteristics. The integration of real-time RGB-D data also presented difficulties, due to noise in the depth maps caused by the low quality of the end-effector's depth sensor and the need for resizing the segmentation mask, which occasionally compromised pre-processing consistency. These challenges highlight the need to optimize pipeline robustness against environmental and object variations, as well as to enhance perception and execution accuracy. To address these limitations, the following future steps are proposed. Firstly, model training should be expanded to include a more diverse dataset, encompassing objects with different shapes, sizes, and textures, along with data augmentation techniques to improve generalization. Secondly, the implementation of smoothing filters or dedicated ML models for depth map denoising can mitigate the impacts of noise, enhancing the quality of inputs for d2nt, pre-processing, and the conversion of the selected pixel to 3D coordinates. 

Furthermore, future work will explore locomotion and manipulation strategies that go beyond the tools provided by the robot's SDK, increasing robustness in dynamic scenarios. Finally, testing in more complex environments, with multiple objects and irregular surfaces, is essential to validate pipeline scalability. These research directions will not only address current limitations but also position deep learning-based loco-manipulation as a viable solution for practical applications in quadruped robotics, such as rescue, logistics, and interaction in domestic environments.

% \section{REFERENCES} \label{sec:references}

\bibliographystyle{IEEEtran}
\bibliography{ref}

@misc{kerasmobilenet,
  author       = {{Keras Team}},
  year         = {2018},
  title        = {Keras Applications: MobileNet},
  howpublished = {\url{https://keras.io/api/applications/mobilenet/}},
  note         = {Acessado em: 07-06-2025}
}

@misc{ronneberger2015unetconvolutionalnetworksbiomedical,
      title={U-Net: Convolutional Networks for Biomedical Image Segmentation}, 
      author={Olaf Ronneberger and Philipp Fischer and Thomas Brox},
      year={2015},
      eprint={1505.04597},
      archivePrefix={arXiv},
      primaryClass={cs.CV},
      url={https://arxiv.org/abs/1505.04597}, 
}

@misc{Free3DWaterBottle,
  author    = {{Bart (3dpixel\_be)}},
  title     = {Bottled Water 3D Model},
  howpublished = {\url{https://free3d.com/3d-model/bottled-water-34022.html}},
  year      = {2025},
  note      = {Accessed: 2025-05-31}
}

@misc{Genesis,
  author    = {{Genesis Project Contributors}},
  title     = {Genesis Framework Documentation},
  howpublished = {\url{https://genesis-world.readthedocs.io/en/latest/index.html}},
  year      = {2025},
  note      = {Accessed: 2025-05-31}
}

@article{Liu2024Overview,
  author    = {Liu, Yang and Li, Bin and Wang, Hesheng and Ge, Shuzhi Sam and Lee, Tong Heng},
  title     = {An Overview of Quadruped Robots: Design, Control, Perception, and Applications},
  journal   = {Applied Sciences},
  volume    = {14},
  number    = {5},
  pages     = {57},
  year      = {2024},
  doi       = {10.3390/app14050057},
}

@article{Papadopoulos2023Legged,
  author    = {Papadopoulos, Evangelos and Kamedula, M. and Vitzilaios, N. and Gounaris, G. and Koutsoukis, K. and Bechlioulis, C. P. and Kostavelis, Ioannis and Giitsidis, Thomas and Tsalatsanis, A. and Tsoli, A. and Amanatiadis, A. and Gasteratos, Antonios and Trahanias, P.},
  title     = {Legged Robot Manipulation: A Review},
  journal   = {Frontiers in Robotics and AI},
  volume    = {10},
  year      = {2023},
  month     = {April},
  pages     = {1142421},
  doi       = {10.3389/fmech.2023.1142421},
}

@inproceedings{Solmaz2024Robust,
  author    = {Solmaz, S. and Innerwinkler, Pamela and Wójcik, Michał and Tong, Kailin and Politi, Elena and Dimitrakopoulos, George and Purucker, Patrick and Höß, Alfred and Schuller, Björn W. and John, Reiner},
  title     = {Robust Robotic Search and Rescue in Harsh Environments: An Example and Open Challenges},
  booktitle = {2024 IEEE International Symposium on Robotic and Sensors Environments (ROSE)},
  year      = {2024},
  month     = {June},
  doi       = {10.1109/ROSE62198.2024.10591144}
}

@article{MachadoSilvaLeggedRobots,
  author    = {Machado, J. A. Tenreiro and Silva, Manuel F.},
  title     = {An Overview of Legged Robots},
  year      = {2006}
}

@article{Guo2024FrameworkGrasp,
  author    = {Guo, Jiamin and Chai, Hui and Zhang, Qin and Zhao, Haoning and Chen, Meiyi and Li, Yueyang and Li, Yibin},
  title     = {A Framework of Grasp Detection and Operation for Quadruped Robot with a Manipulator},
  journal   = {Drones},
  volume    = {8},
  number    = {5},
  pages     = {208},
  year      = {2024},
  month     = {May},
  doi       = {10.3390/drones8050208}
}

@phdthesis{Mahler2018DeepLearningGrasping,
  author    = {Mahler, Jeffrey},
  title     = {Deep Learning for Robust Robot Grasping from Synthetic Data},
  school    = {University of California, Berkeley},
  year      = {2018},
  month     = {August},
  note      = {Technical Report No. UCB/EECS-2018-120}
}

@misc{kasaei2022simultaneousmultiviewobjectrecognition,
      title={Simultaneous Multi-View Object Recognition and Grasping in Open-Ended Domains}, 
      author={Hamidreza Kasaei and Sha Luo and Remo Sasso and Mohammadreza Kasaei},
      year={2022},
      eprint={2106.01866},
      archivePrefix={arXiv},
      primaryClass={cs.RO}
}

@misc{morrisonclosing2018,
	title = {Closing the Loop for Robotic Grasping: A Real-time, Generative Grasp Synthesis Approach},
	doi = {10.48550/arXiv.1804.05172},
	shorttitle = {Closing the Loop for Robotic Grasping},
	number = {{arXiv}:1804.05172},
	publisher = {{arXiv}},
	author = {Morrison, Douglas and Corke, Peter and Leitner, Jürgen},
	date = {2018-05-15},
        year = {2018},
	eprinttype = {arxiv},
	eprint = {1804.05172 [cs]},
}

@software{yolo11_ultralytics,
  author = {Glenn Jocher and Jing Qiu},
  title = {Ultralytics YOLO11},
  version = {11.0.0},
  year = {2024},
  url = {https://github.com/ultralytics/ultralytics},
  orcid = {0000-0001-5950-6979, 0000-0002-7603-6750, 0000-0003-3783-7069},
  license = {AGPL-3.0}
}

@misc{BostonDynamicsSpotSDK,
  author    = {{Boston Dynamics}},
  title     = {{Spot SDK}: Software Development Kit for the Spot Robot},
  howpublished = {\url{https://github.com/boston-dynamics/spot-sdk}},
  year      = {2025},
  note      = {Accessed: 2025-05-31}
}

%%%%%%%%%%%%%%%%%%%%%%%%%%%%%%%%%%%%%%%%%%%%%%%%%%%%%%%%%%%%%%%%%%%%%%%%%%%%%%%%
%\section*{APPENDIX}

%This work was supported by Fundação de Apoio à Física e à Química (FAFQ) and funded by Petrobras and the Federal Government of Brazil, grant no. ANP-Petrobras 2023/00013-7 and 2023/00016-6.

\end{document}